\documentclass[letterpaper,10pt,conference]{ieeeconf}

\IEEEoverridecommandlockouts
\usepackage{amsmath,amssymb}
\usepackage{graphicx}
\usepackage{booktabs}
\usepackage{array}
\usepackage{algorithm}
\usepackage{algpseudocode}
\usepackage{xcolor}
\usepackage{cite}
\usepackage{url}
\usepackage{balance}
\usepackage{needspace}

\graphicspath{{figures/}}
\newcommand{\method}{Outcome-Sensitive Motion Search}
\newcommand{\clip}{\operatorname{clip}}
\algrenewcommand\algorithmicrequire{\textbf{Input:}}
\algrenewcommand\algorithmicensure{\textbf{Output:}}
\algtext*{EndIf}
\algtext*{EndFor}

\usepackage{xcolor}

\title{\LARGE \bf
Outcome-Sensitive Motion Search for Impact-Aware Dexterous Catching
}

\author{
Guorui Pei$^{1}$, Jinsong Wu$^{2}$, Songyuan Su$^{3,4}$, Jiaming Qi$^{5}$,\\
Sichao Liu$^{6}$, David Navarro-Alarcon$^{2}$, Bin Liu$^{7,*}$, and Peng Zhou$^{4,*}$\\[1mm]
{\small $^{1}$College of Robotics Science and Engineering, Taiyuan University of Technology, Taiyuan, China}\\
{\small $^{2}$Department of Mechanical Engineering, The Hong Kong Polytechnic University, Kowloon, Hong Kong SAR, China}\\
{\small $^{3}$Southern University of Science and Technology, Shenzhen, China}\\
{\small $^{4}$School of Advanced Engineering, Great Bay University, Dongguan, Guangdong, China}\\
{\small $^{5}$College of Mechanical and Electrical Engineering, Northeast Forestry University, Harbin, China}\\
{\small $^{6}$Department of Production Engineering, KTH Royal Institute of Technology, Stockholm, Sweden}\\
{\small $^{7}$Kaiyang Laboratory, Chery Automobile Co., Ltd., Wuhu, Anhui, China}\\
{\small $^{*}$Corresponding authors.}
}

\begin{document}
\maketitle
\thispagestyle{empty}
\pagestyle{empty}

\begin{abstract}
Skilled humans can catch fast-moving objects softly by coordinating interception, velocity matching, and follow-through to mitigate impact. Learning such impact-aware catching with reinforcement learning (RL), however, is challenging, as the policy must achieve reliable interception and grasping while regulating the sensitive transition into contact. Moreover, even a capable privileged-state RL teacher may not provide ideal demonstrations for a deployable imitation-learning (IL) student: teacher failures limit task coverage, while small variations in pre-contact motion can produce substantially different impact and grasping outcomes. We characterize this phenomenon through interventional outcome sensitivity and introduce the outcome-sensitive window (OSW) to guide targeted demonstration construction. Building on this formulation, we propose Outcome-Sensitive Motion Search, which learns a task-conditioned manifold of successful OSW motions and performs local geodesic search to refine successful teacher rollouts and repair task conditions where the teacher fails. We then validate candidate motions through complete rollouts under a calibrated IL-student action-error model and retain only successful executions as demonstrations. Extensive simulation experiments demonstrate that our method effectively repairs task conditions where the teacher fails and enables the resulting IL policy to outperform the privileged RL teacher in both catching success and impact mitigation.
\end{abstract}
\section{Introduction}

Skilled humans can catch fast-moving objects softly by coordinating interception, velocity matching, and follow-through to mitigate impact. Reproducing this capability on robots is challenging because successful catching requires the robot to simultaneously intercept the object, regulate the transition into contact, and maintain a stable grasp~\cite{salehian2016softly,zhao2023impactfriendly,yan2024impact,tassi2026imacatcher}. Reinforcement learning (RL) provides a promising approach for acquiring such coordinated behaviors~\cite{hu2023modular,lan2024dexcatch,zhang2025catchit}, but impact-aware catching remains difficult because pre-contact motion, contact dynamics, and post-contact grasping are tightly coupled.

Privileged-state RL can learn effective catching behaviors in simulation, yet a capable RL teacher does not necessarily provide ideal demonstrations for a deployable imitation-learning (IL) student. Teacher failures limit the coverage of success-only demonstrations, while successful rollouts can vary considerably in impact quality and robustness to execution errors~\cite{mandlekar2022offline,chen2025s2i}. More importantly, the influence of an action on the final outcome is highly nonuniform along a catching trajectory~\cite{wang2026world}: small variations shortly before contact can substantially alter interception, impact, and subsequent grasping. We characterize this property through \emph{interventional outcome sensitivity} and introduce an \emph{outcome-sensitive window} (OSW), instantiated for catching as the short pre-contact interval where local motion interventions can strongly affect downstream outcomes.

Building on this insight, we propose \emph{Outcome-Sensitive Motion Search} for constructing improved demonstrations from mixed teacher experience. We learn a task-conditioned manifold of successful OSW motions and perform local geodesic search to generate alternative pre-contact motions. The search serves two complementary purposes: \emph{refinement} improves successful teacher rollouts, particularly their impact quality, while \emph{repair} recovers successful executions for task conditions where the teacher fails. Each candidate is evaluated through a complete simulator rollout to account for its downstream contact and grasping consequences.

\begin{figure}[t]
  \centering
  \IfFileExists{figures/first_figure.pdf}{    \includegraphics[width=0.8\columnwidth]{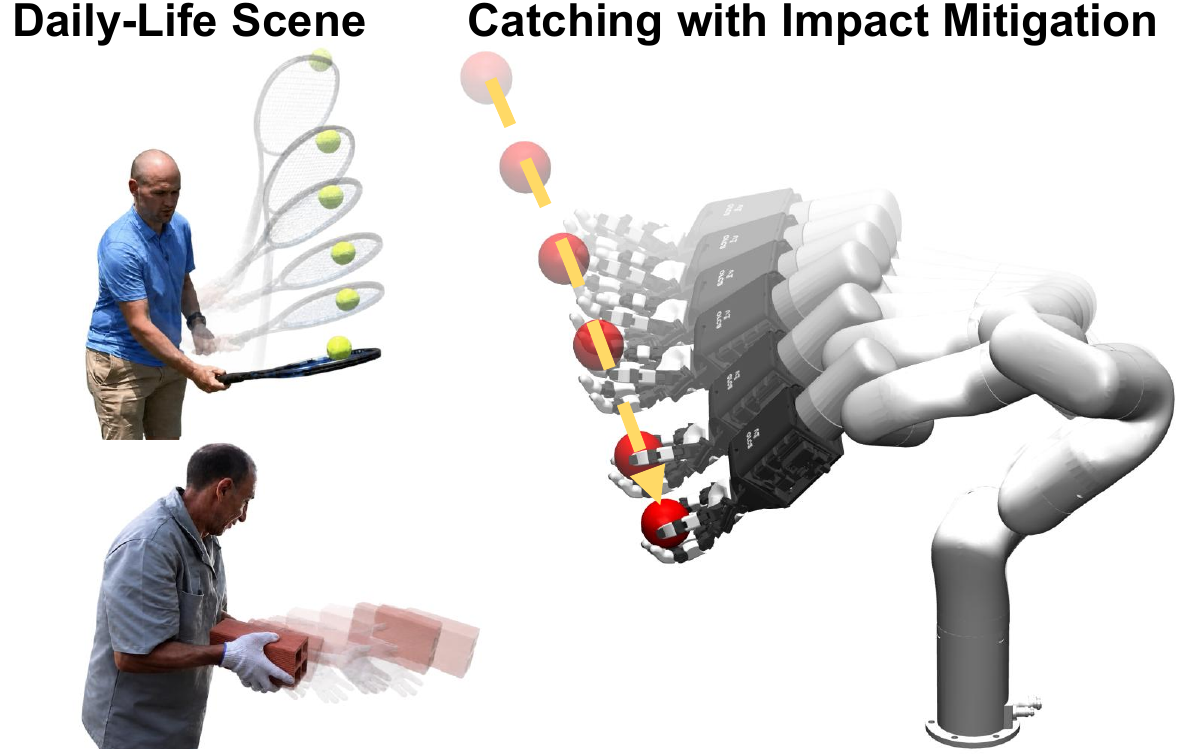}  }{    \fbox{\parbox[c][1.72in][c]{0.5\columnwidth}{\centering\footnotesize
    Upload \texttt{first\_figure.pdf} to the \texttt{figures/} directory.}}  }

  \vspace{0mm}   \caption{Motivation for impact-aware dexterous catching. Human and robot examples illustrate interception, velocity matching, and follow-through for impact mitigation.}
  \label{fig:motivation}
\vspace{0mm} \end{figure}

Successful execution alone, however, does not guarantee that a demonstration is suitable for an IL student operating with restricted observations~\cite{ross2011dagger,laskey2017dart}. We therefore learn an IL-student action-error model from IL students and use it to evaluate candidate motions through complete rollouts under modeled student execution errors. Only qualified executions are retained for final IL student training. Extensive simulation experiments demonstrate that the resulting IL policy surpasses the privileged RL teacher in both catching success and impact mitigation, with ablations confirming the contributions of motion refinement and repair, manifold-based search, and student-aware validation.

The main contributions of this work are:
\begin{itemize}
    \item We introduce an outcome-sensitive formulation for demonstration construction and instantiate the OSW for dynamic catching.
    \item We propose OSW motion manifold learning and local geodesic search to refine successful teacher motions and repair failed task conditions.
    \item We develop student-aware complete-rollout validation using a calibrated IL-student action-error model, and extensively evaluate the framework across diverse catching conditions.
\end{itemize}

\section{Related Work}

\subsection{Impact-Aware Manipulation}

Impact-mitigating catching is an important problem in dynamic manipulation. Model-based approaches use trajectory optimization and model predictive control to determine interception timing, catching configurations, and robot motion
~\cite{bauml2010catching,lampariello2011trajectory,gold2022mpc}, while soft-catching and impact-aware controllers regulate robot--object relative motion through compliant, variable-stiffness, or force control
~\cite{salehian2016softly,zhao2023impactfriendly,yan2024impact,tassi2026imacatcher}.
These methods rely on accurate state and dynamics models and may incur substantial online computation.

Learning-based methods provide an alternative for high-dimensional arm--hand coordination. Reinforcement and imitation learning have enabled dynamic catching across different objects and throwing conditions
~\cite{hu2023modular,lan2024dexcatch,zhang2025catchit}, with recent work further combining learning and structured impact-aware motion generation ~\cite{pei2026kinodynamic,tassi2026imacatcher}.
However, learned policies can remain imperfect in both catching success and impact mitigation. Instead of treating such a policy as the final solution, we use it as a teacher and further improve its experience before training a final student.

\subsection{Policy Learning from Imperfect Data}

Imitation learning depends strongly on demonstration quality and state coverage
~\cite{mandlekar2022offline}.
Existing methods improve mixed-quality data by filtering, reweighting, or selecting demonstrations
~\cite{xu2022dwbc,kim2022demodice}, or by augmenting demonstrations, collecting corrective or interventional data, and refining suboptimal trajectories
~\cite{mandlekar2023mimicgen,ke2024ccil,hoque2024intervengen,hertel2021learning,chen2025s2i}.
These approaches improve the supervision available to the learner, but demonstration quality under unperturbed execution alone does not guarantee effective closed-loop imitation.

Small prediction errors can drive an imitation policy away from the demonstrated state distribution and compound over time
~\cite{ross2011dagger,laskey2017dart}.
DAgger addresses this problem by collecting learner-induced states, whereas DART perturbs expert demonstrations to expose the learner to states arising from execution errors
~\cite{ross2011dagger,laskey2017dart}.
For impact-mitigating catching, where small action deviations near contact can substantially affect downstream outcomes, demonstration improvement and learner robustness must therefore be considered jointly. Our work addresses both by improving teacher experience in catching success and impact mitigation while validating the resulting motions under modeled learner action errors.

\begin{figure}[t]
  \centering
  \IfFileExists{figures/problem_formulationA.pdf}{    \includegraphics[width=0.9\columnwidth]{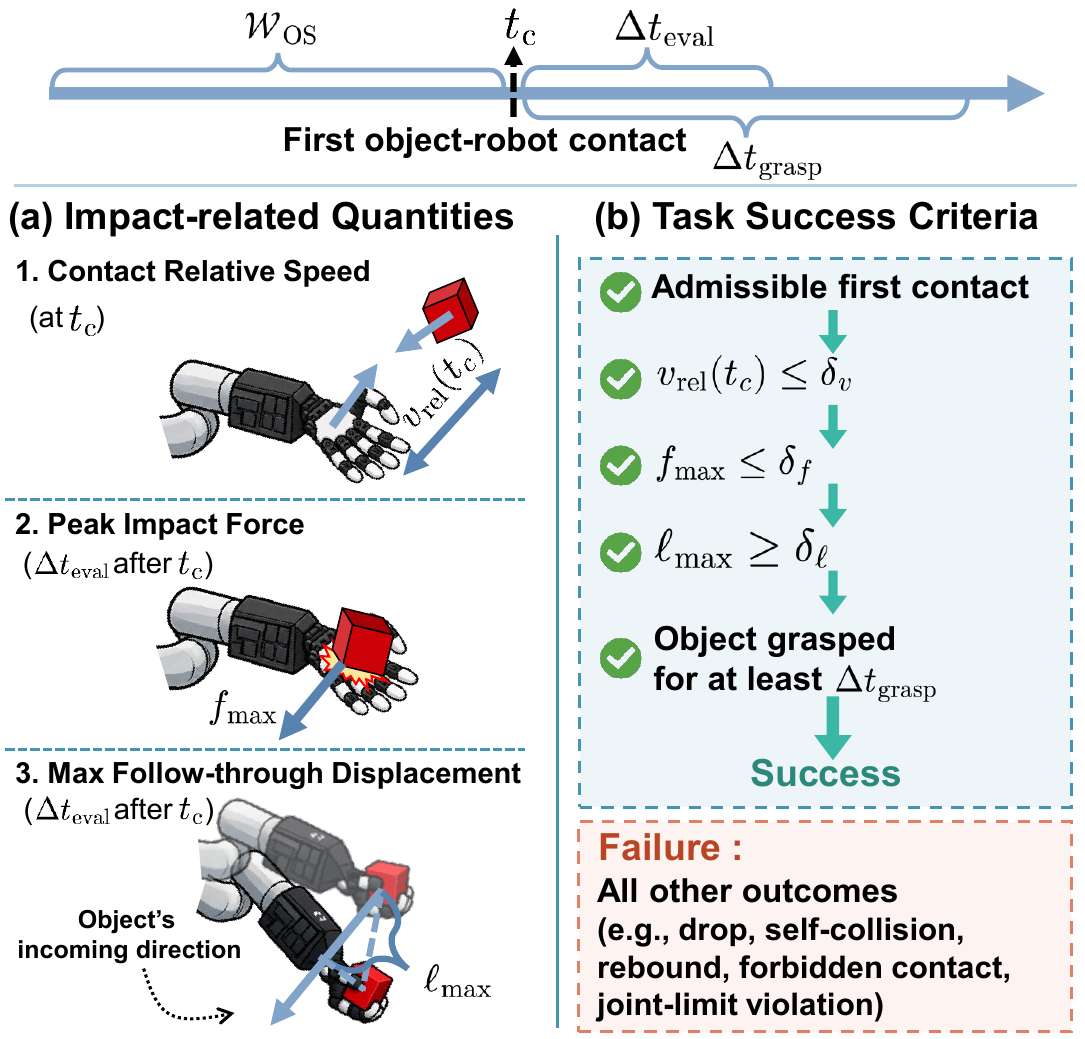}  }{    \fbox{\parbox[c][2.70in][c]{0.94\columnwidth}{\centering\footnotesize
    Upload \texttt{problem\_formulationA.pdf} to the \texttt{figures/} directory.}}  }
  \vspace{-2mm}   \caption{Task quantities and success criteria: (a) contact relative speed,
  peak impact force, and maximum follow-through displacement; (b) conditions for
  task success.}
  \label{fig:problemformulation}
\vspace{-1mm} \end{figure}

\begin{figure*}[t]
  \centering
  \includegraphics[width=0.98\textwidth]{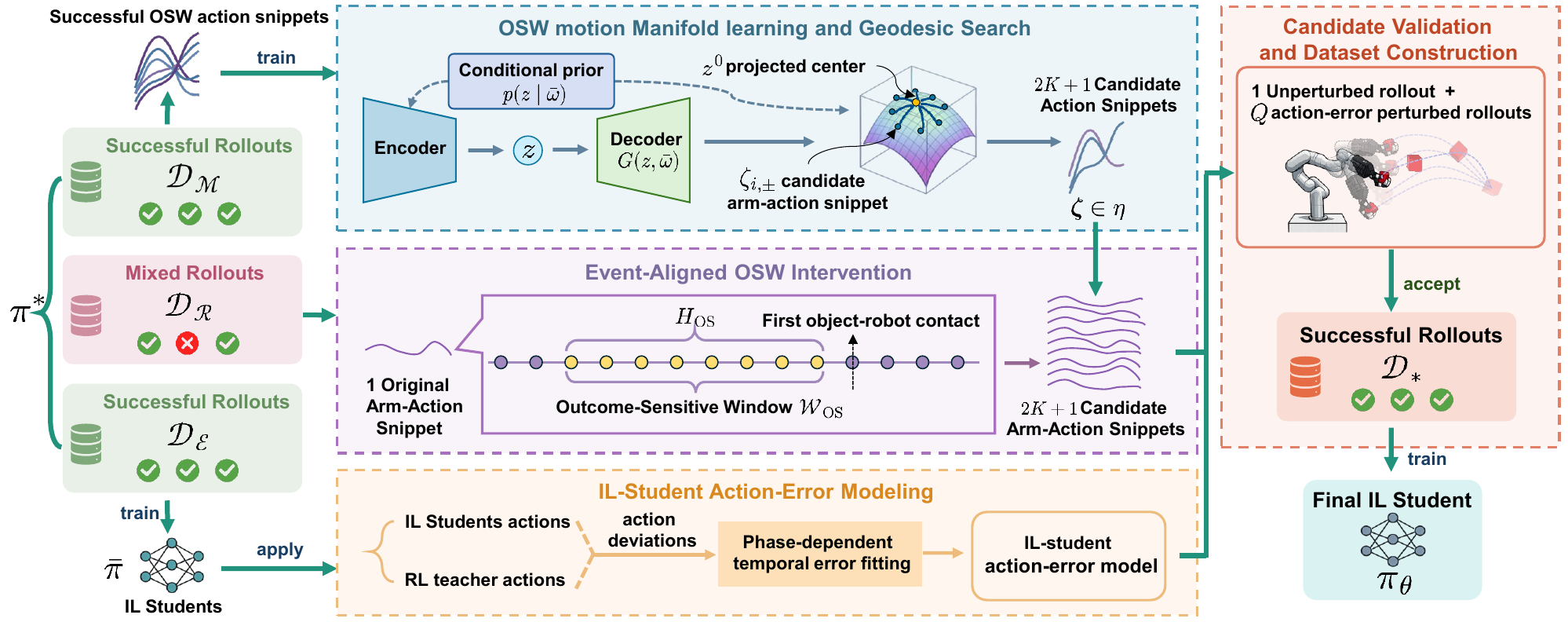}
  \vspace{-2mm}   \caption{Overview of \method. RL experience flows through event alignment,
  geodesic search on the OSW motion manifold, student-aware complete-rollout
  validation, and construction of the final IL training dataset.}
  \label{fig:framework}
\vspace{-2mm} \end{figure*}

\section{Problem Formulation}
\label{sec:problem}

\subsection{Task and Success Criteria}

We study impact-aware dexterous catching using a $n_a$-DoF arm and a $n_h$-DoF hand.  A frozen RL teacher $\pi^{*}$ has access to privileged simulator states, whereas the final student policy $\pi_{\theta}$ receives only a history of deployable observations and previous actions. Both policies output $n$-dimensional arm--hand actions, with each action channel normalized to $[-1,1]$. A task condition $\omega \sim p(\omega)$ specifies the initial object state, robot state, and environment parameters.

We characterize catching performance using three quantities. Let $t_c$ denote the time of the first admissible hand--object contact. The contact relative speed $v_{\mathrm{rel}}(t_c)$ is the magnitude of the object--palm relative velocity immediately before $t_c$. The peak impact force $f_{\mathrm{max}}$ is the maximum summed normal force over admissible contacts within an evaluation interval $\Delta t_{\mathrm{eval}}$ after $t_c$. The maximum follow-through displacement \(\ell_{\max}\) is the maximum palm displacement along the pre-contact incoming direction over the same interval.
A rollout is successful if the first object--robot contact is an admissible hand--object contact, $v_{\mathrm{rel}}(t_c)\leq\delta_v$, $f_{\mathrm{max}}\leq\delta_f$, $\ell_{\mathrm{max}}\geq\delta_{\ell}$, and the object remains grasped for at least $\Delta t_{\mathrm{grasp}}$, where $\Delta t_{\mathrm{grasp}}\gg\Delta t_{\mathrm{eval}}$. Forbidden contact, joint-limit violation, self-collision, rebound, drop, or violation of any success criterion constitutes failure. The numerical values are given in Section~\ref{sec:systemsetup} and summarized in Fig.~\ref{fig:problemformulation}. For a complete rollout $\tau=(s_0,a_0,s_1,a_1,\ldots,s_T)$, we denote task success by $\mathbb{I}_{\mathrm{succ}}(\tau)\in\{0,1\}$.

\subsection{Interventional Outcome Sensitivity}
Dynamic manipulation can contain short temporal intervals in which local changes in robot motion produce disproportionately large variations in downstream task outcomes. We characterize this property through \emph{interventional outcome sensitivity} (IOS). Consider a temporal window $\mathcal{W}$ of a rollout and its corresponding action snippet $\boldsymbol{\zeta}_{\mathcal W}=\{a_k\}_{k\in\mathcal W}$. Under the same task condition $\omega$, replacing $\boldsymbol{\zeta}_{\mathcal W}$ by a locally perturbed segment $\boldsymbol{\zeta}_{\mathcal W}+\Delta\boldsymbol{\zeta}$ constitutes an intervention whose effect is measured through the resulting downstream task outcome $Y$.

For a window $\mathcal W$, we conceptually define its interventional outcome sensitivity as
\begin{equation}
\mathcal S_{\mathrm{IOS}}(\mathcal W;\omega)
=
\mathbb E_{\Delta\boldsymbol{\zeta}}
\left[
\frac{
d_Y\!\left(
Y_\omega(\boldsymbol{\zeta}_{\mathcal W}),
Y_\omega(\boldsymbol{\zeta}_{\mathcal W}+\Delta\boldsymbol{\zeta})
\right)}
{\|\Delta\boldsymbol{\zeta}\|}
\right],
\label{eq:ios}
\end{equation}
where $d_Y(\cdot,\cdot)$ measures the variation between downstream outcomes induced by a local action intervention. A temporal segment exhibiting high IOS is termed an \emph{outcome-sensitive window} (OSW), denoted by $\mathcal W_{\mathrm{OS}}$.

For dynamic catching, we align the OSW to a rollout-specific interception event. Let \(k_e\) denote the control step of the first object--robot contact if contact occurs; otherwise, \(k_e\) is the step of closest approach between the object and palm. We instantiate the OSW as the \(H_{\mathrm{OS}}\) control steps
immediately preceding this event, with the associated OSW motion
defined as
\[
\begin{aligned}
\mathcal W_{\mathrm{OS}}
&=\{k_e-H_{\mathrm{OS}},\ldots,k_e-1\},\\
\boldsymbol{\zeta}_{\mathrm{OS}}
&=(a_{k_e-H_{\mathrm{OS}}},\ldots,a_{k_e-1}).
\end{aligned}
\]

For successful rollouts, the alignment event coincides with the first admissible hand--object contact. Our demonstration construction modifies only $\boldsymbol{\zeta}_{\mathrm{OS}}$ while retaining the teacher policy outside the OSW. This localizes search to the portion of the trajectory whose motion has the strongest downstream influence while preserving the remainder of the teacher behavior.

\subsection{Outcome-Sensitive Demonstration Construction}

Given teacher experience collected over task conditions $\omega\sim p(\omega)$, our objective is to construct a set of successful demonstrations that improves the closed-loop performance of an observation-restricted student.

For a task condition where the teacher succeeds, we seek an alternative OSW motion $\boldsymbol{\zeta}_{\mathrm{OS}}'$ that preserves task success while reducing impact and remaining robust to modeled student action errors. For a task condition where the teacher fails, we instead seek a replacement OSW motion that converts the rollout into a successful and robust demonstration. In both cases, actions outside $\mathcal W_{\mathrm{OS}}$ remain governed by the frozen teacher.

Let $\tau(\omega,\boldsymbol{\zeta}_{\mathrm{OS}})$ denote the complete rollout obtained under task condition $\omega$ when the OSW motion is replaced by $\boldsymbol{\zeta}_{\mathrm{OS}}$. We seek
\begin{equation}
\boldsymbol{\zeta}_{\mathrm{OS}}^{*}
=
\arg\min_{\boldsymbol{\zeta}\in\mathcal M_{\mathrm{OS}}(\omega)}
J\!\left(\tau(\omega,\boldsymbol{\zeta})\right),
\label{eq:demo_construction}
\end{equation}
subject to
\begin{equation}
\mathbb I_{\mathrm{succ}}\!\left(\tau(\omega,\boldsymbol{\zeta})\right)=1,
\qquad
\mathrm{SR}(\boldsymbol{\zeta}\mid\omega)\geq\delta_{\mathrm{SR}},
\label{eq:demo_constraints}
\end{equation}
where $\mathcal M_{\mathrm{OS}}(\omega)$ denotes the OSW motion manifold, $J$ evaluates demonstration quality, and $\mathrm{SR}(\cdot)$ measures robustness under modeled student action errors. Only complete rollouts satisfying the task-success criteria under their original task conditions are retained as positive demonstrations for student learning.

\section{Outcome-Sensitive Motion Search}
\label{sec:method}
Our framework searches candidate arm-action snippets within the outcome-sensitive window (OSW) and validates the resulting complete rollouts under modeled student action errors, as illustrated in Fig.~\ref{fig:framework}. For dynamic catching, the OSW corresponds to the short pre-contact interval defined in Section~\ref{sec:problem}, and we restrict the intervention within this window to the arm actions, whose variations can strongly affect the subsequent impact and grasping outcomes. We construct an OSW motion manifold from successful teacher experience and perform local geodesic search on this manifold to generate candidate replacement snippets.

Three disjoint datasets, collected i.i.d. from the same frozen RL teacher $\pi^{*}$ under task conditions $\omega\sim p(\omega)$, serve distinct roles. $\mathcal D_{\mathcal M}$ contains successful rollouts for learning the OSW motion manifold, $\mathcal D_{\mathcal E}$ contains successful rollouts for calibrating the IL-student action-error model, and $\mathcal D_{\mathcal R}$ contains mixed successful and failed rollouts whose OSW motions are respectively refined or repaired. Candidate motions are evaluated through complete simulator rollouts under unperturbed execution and calibrated student action errors, and only successful executions are retained in the constructed demonstration dataset $\mathcal D_{*}$. Privileged state and contact information are used only during this offline construction process; $\mathcal D_{*}$ contains only deployable observation histories and executed-action labels for training the final student $\pi_{\theta}$.

\subsection{OSW Motion Manifold Learning and Geodesic Search}
\label{sec:manifold}

The OSW motion manifold represents successful arm-action snippets within the outcome-sensitive window in a low-dimensional latent space. Let $\boldsymbol{\zeta}_{\mathrm{OS}}$ denote an OSW action snippet. A conditional decoder $G(z,\bar{\omega})$ maps a latent code $z$ to $\boldsymbol{\zeta}_{\mathrm{OS}}$, where the conditioning variable $\bar{\omega}$ augments the task condition $\omega$ with the privileged simulator state at the beginning of the OSW.
We learn the OSW motion manifold from successful rollouts in $\mathcal D_{\mathcal M}$ using a conditional variational model. A diagonal-Gaussian encoder maps each successful OSW action snippet and its condition to a latent distribution, while a conditional prior $p(z\mid\bar{\omega})$ models likely latent codes under that condition. The decoder uses a $\tanh$ output layer to enforce the normalized action bounds. Training minimizes a weighted combination of reconstruction, temporal-smoothing, and Kullback--Leibler divergence losses. Reconstruction is measured using the weighted Euclidean norm $\|\cdot\|_{W_a}$, where the positive-definite matrix $W_a$ normalizes the action channels. The smoothing term penalizes squared temporal second differences of the decoded actions, while the KL term regularizes the encoded distribution toward the conditional prior.

For each rollout in $\mathcal D_{\mathcal R}$, we first project its original OSW action snippet $\boldsymbol{\zeta}_{\mathrm{org}}$ onto the learned manifold. With $\bar{\omega}$ fixed, the projected latent center is obtained over the bounded latent region $\mathcal Z$ as
\begin{equation}
\begin{aligned}
z^0\in\arg\min_{z\in\mathcal Z}\;&
\left\|G(z,\bar{\omega})-\boldsymbol{\zeta}_{\mathrm{org}}\right\|_{W_a}^2\\
&-\lambda_p\log p(z\mid\bar{\omega}),
\end{aligned}
\label{eq:projection}
\end{equation}
where $\lambda_p>0$ discourages projection into low-probability regions of the conditional prior. Samples from the encoder and conditional prior initialize a fixed multi-start search. The resulting $z^0$ serves as the center for local manifold search; its decoded motion $G(z^0,\bar{\omega})$ need not exactly reproduce the original RL snippet.

Because equal displacements in latent space can induce substantially different changes in decoded actions, Euclidean latent distance does not faithfully characterize local motion variation. We therefore equip the latent space with the decoder-induced pullback metric
\begin{equation}
\mathbf M(z,\bar{\omega})
=
J_G^\top W_a J_G+\lambda_{\mathrm{reg}}\mathbf I,
\label{eq:pullback}
\end{equation}
where $J_G$ denotes the decoder Jacobian with respect to $z$, evaluated at $(z,\bar{\omega})$. The metric measures latent displacement according to its induced change in the decoded OSW motion, while $\lambda_{\mathrm{reg}}>0$ keeps $\mathbf M$ nonsingular.
Starting from $z^0$, we generate $K$ tangent directions $\xi_i$, $i=1,\ldots,K$, from a fixed Sobol low-discrepancy sequence and orthogonalize and normalize them under $\mathbf M(z^0,\bar{\omega})$. Let $\gamma_{i,\pm}$ denote geodesics under the pullback metric, with $\bar{\omega}$ fixed and shooting radius $r$:
\begin{equation}
\begin{aligned}
\gamma_{i,\pm}(0)&=z^0,\qquad
\dot{\gamma}_{i,\pm}(0)=\pm r\xi_i,\\
\boldsymbol{\zeta}_{i,\pm}
&=
G\!\left(\gamma_{i,\pm}(1),\bar{\omega}\right).
\end{aligned}
\label{eq:shooting}
\end{equation}
The $2K$ geodesic endpoints together with the projected center
$\boldsymbol{\zeta}_0=G(z^0,\bar{\omega})$ form the candidate set
\begin{equation}
\eta=
\left\{
\boldsymbol{\zeta}_0,
\boldsymbol{\zeta}_{1,+},
\boldsymbol{\zeta}_{1,-},
\ldots,
\boldsymbol{\zeta}_{K,+},
\boldsymbol{\zeta}_{K,-}
\right\},
\qquad |\eta|=2K+1.
\label{eq:candidate_set}
\end{equation}
The shooting radius is bounded so that each geodesic remains within $\mathcal Z$ and the decoded candidates remain local to the projected original motion. Since the manifold is learned from successful OSW motions but does not guarantee that every local candidate remains successful under its target task condition, each candidate is subsequently evaluated through complete simulator rollouts.

\subsection{Event-Aligned OSW Intervention}
\label{sec:precontactrollout}

For dynamic catching, we instantiate the outcome-sensitive window (OSW) as the \(H_{\mathrm{OS}}\) control steps immediately preceding the rollout-specific interception event \(k_e\) defined in Section~\ref{sec:problem}. If \(k_e\) occurs too early to provide a complete \(H_{\mathrm{OS}}\)-step window, search is skipped and the original rollout is retained only if successful.

Let $\boldsymbol{\zeta}_{\mathrm{org}}$ denote the original OSW arm-action snippet and $\boldsymbol{\zeta}\in\eta$ a candidate replacement. Before the OSW, the frozen RL teacher $\pi^{*}$ controls the complete arm--hand system; within the OSW, $\boldsymbol{\zeta}$ is executed open loop for the arm while $\pi^{*}$ continues closed-loop hand control, preserving its learned finger coordination. At the first actual object--robot contact or the end of the OSW, whichever occurs first, $\pi^{*}$ resumes full arm--hand control until termination. Each candidate is thus evaluated by its complete rollout $\tau(\omega,\boldsymbol{\zeta})$ rather than by the modified OSW segment alone.

\subsection{IL-Student Action-Error Modeling}
\label{sec:errormodel}

To model execution errors expected for the final student $\pi_{\theta}$, we train a set of IL students, denoted by $\bar{\pi}$, with the same observation interface, architecture, and training protocol as $\pi_{\theta}$. Each IL student $\bar{\pi}$ is trained on a subset of $\mathcal D_{\mathcal E}$ and evaluated on held-out task conditions through cross-fitting. At states visited by $\bar{\pi}$, its executed actions are paired with actions queried from the frozen RL teacher $\pi^{*}$. We model the resulting $\bar{\pi}$--teacher deviations using phase-dependent action delay, channel-wise gain, bias, and temporally correlated residuals, where the catching phase $\varphi$ (approach, velocity matching, contact, follow-through, or grasping) is determined online from object--palm proximity and contact history.

At control step $n$, let $\bar a_n$ denote the intended arm--hand action under the OSW intervention protocol. The modeled action of $\pi_{\theta}$, $\widetilde a_n$, and perturbed action $a_n$ used for candidate validation are
\begin{equation}
\begin{aligned}
e_n &= A_\varphi e_{n-1}+L_\varphi\nu_n,\\
\widetilde a_n &= S_\varphi\bar a_{n-d_\varphi}+\mu_\varphi+e_n,\\
a_n &= \clip\!\left(
\bar a_n+\kappa_\epsilon[\widetilde a_n-\bar a_n],-1,1
\right),
\end{aligned}
\label{eq:errormodel}
\end{equation}
where $d_\varphi$, diagonal $S_\varphi$, and $\mu_\varphi$ capture systematic deviations, while $e_n$ follows a stable autoregressive process driven by $\nu_n\sim\mathcal N(0,\mathbf I_n)$. The factor $\kappa_\epsilon>1$ scales the modeled deviation. We select $d_\varphi$ over a bounded grid, fit $S_\varphi$ and $\mu_\varphi$ by least squares, and estimate $A_\varphi$ and $L_\varphi$ from the resulting residual dynamics and covariance. During validation, $e_n$ is initialized from the fitted stationary distribution and the error model is applied to all action channels throughout the complete rollout.

\subsection{Candidate Validation and Dataset Construction}
\label{sec:selection}

Each candidate OSW action snippet $\boldsymbol{\zeta}\in\eta$ is evaluated through one unperturbed rollout $\tau^{(0)}(\omega,\boldsymbol{\zeta})$ and $Q$ action-error-perturbed rollouts $\tau^{(q)}(\omega,\boldsymbol{\zeta})$ under the calibrated action-error model fitted from $\bar{\pi}$ to approximate execution errors of $\pi_{\theta}$. Its perturbed success rate is
\begin{equation}
\mathrm{SR}(\boldsymbol{\zeta}\mid\omega)
=
\frac{1}{Q}\sum_{q=1}^{Q}
\mathbb I_{\mathrm{succ}}
\!\left(\tau^{(q)}(\omega,\boldsymbol{\zeta})\right),
\label{eq:perturbed_sr}
\end{equation}
which measures robustness to modeled execution errors of $\pi_{\theta}$. Within each task condition, all candidates share the same initial residuals and Gaussian innovations for fair comparison. For a candidate that succeeds under unperturbed execution, we define
\begin{equation}
J(\boldsymbol{\zeta})
=
\lambda_f f_{\mathrm{max}}
+
\lambda_d
\left\|
\boldsymbol{\zeta}-\boldsymbol{\zeta}_{\mathrm{org}}
\right\|_{W_a},
\label{eq:unperturbedquality}
\end{equation}
where $\lambda_f,\lambda_d>0$ balance impact mitigation and deviation from the original OSW motion.

Eligibility depends on the original RL outcome. Accordingly, the generic
robustness constraint in Eq.~\eqref{eq:demo_constraints} is instantiated
relative to the original rollout for refinement and as an absolute
threshold for repair. For \emph{refinement} of a successful rollout,
$\eta_{\mathrm{elig}}$ contains candidates that succeed under unperturbed
execution, satisfy $J(\boldsymbol{\zeta})<J_{\mathrm{org}}$, and achieve
$\mathrm{SR}(\boldsymbol{\zeta}\mid\omega)\geq
\mathrm{SR}_{\mathrm{org}}$. For \emph{repair} of a failed rollout, where
no successful original provides a robustness reference,
$\eta_{\mathrm{elig}}$ contains candidates that succeed under unperturbed
execution and satisfy
$\mathrm{SR}(\boldsymbol{\zeta}\mid\omega)\geq\delta_{\mathrm{SR}}$.
From $\eta_{\mathrm{elig}}$, we select the candidate with the lowest $J$,
breaking ties by higher $\mathrm{SR}$ and then shorter geodesic distance
from $z^0$.If no candidate qualifies, a successful RL original is retained, whereas a failed condition is discarded.

\begin{figure}[htbp]
  \centering
  \includegraphics[width=\columnwidth]{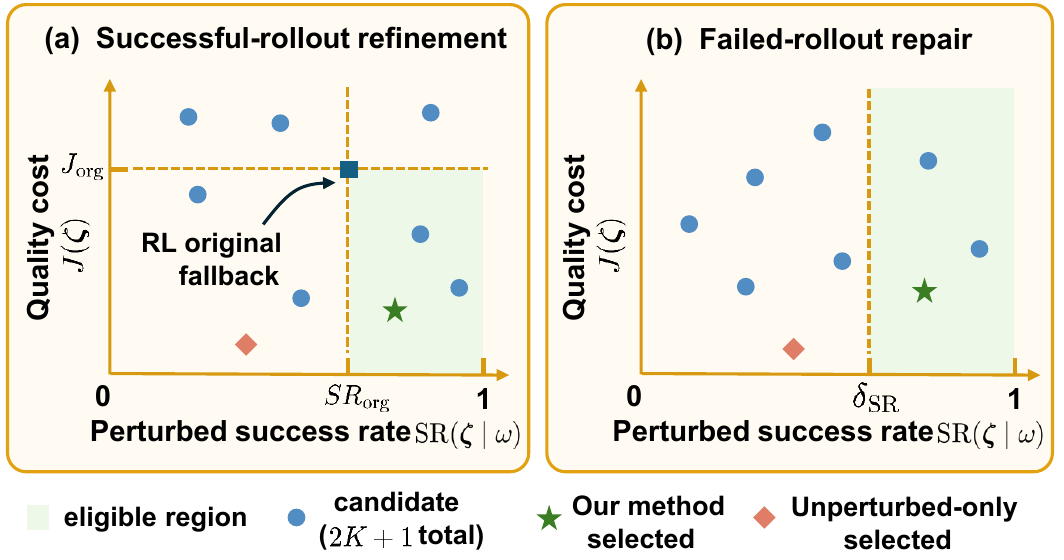}
  \vspace{-7mm} 
  \caption{Student-aware candidate selection for (a) successful-rollout
  refinement and (b) repair. 
  }
  \label{fig:candidateselection}
\vspace{-2mm}
\end{figure}

For the selected $\boldsymbol{\zeta}^{*}$, its unperturbed rollout and all successful action-error-perturbed rollouts are retained. Together with retained successful RL originals, they form $\mathcal D_{*}$. The final student $\pi_{\theta}$ receives only deployable observation histories and previous actions, while labels are the actions actually executed in the retained rollouts, including modeled perturbations. Consecutive labels are grouped into fixed-length action chunks. Alg.~\ref{alg:osms} summarizes the complete construction procedure.

\begin{algorithm}[htbp]
\caption{Outcome-sensitive motion search for demonstration construction}
\label{alg:osms}
\footnotesize
\begin{algorithmic}[1]
\Require $\pi^{*}$; $\mathcal D_{\mathcal M},\mathcal D_{\mathcal E},\mathcal D_{\mathcal R}$;
$H_{\mathrm{OS}},Q,\delta_{\mathrm{SR}}$
\Ensure $\mathcal D_{*}$

\Statex \textcolor{blue}{\textit{1. Model fitting}}
\State Learn the OSW motion manifold from $\mathcal D_{\mathcal M}$.
\State Fit the $\bar{\pi}$-based action-error model on $\mathcal D_{\mathcal E}$.
\Comment{\textcolor{blue}{\eqref{eq:errormodel}}}

\ForAll{$(\omega,\tau_{\mathrm{org}})\in\mathcal D_{\mathcal R}$}

  \Statex \hspace{\algorithmicindent}\textcolor{blue}{\textit{2. OSW motion search}}
  \State Align the outcome-sensitive window $\mathcal W_{\mathrm{OS}}$.
  \If{$\mathcal W_{\mathrm{OS}}$ is incomplete}
    \State Retain $\tau_{\mathrm{org}}$ if
    $\mathbb I_{\mathrm{succ}}(\tau_{\mathrm{org}})=1$; \textbf{continue}.
  \EndIf
  \State Extract the original OSW arm-action snippet
  $\boldsymbol{\zeta}_{\mathrm{org}}$.
  \State Project $\boldsymbol{\zeta}_{\mathrm{org}}$ to $z^0$.
  \Comment{\textcolor{blue}{\eqref{eq:projection}}}
  \State Generate candidate set $\eta$ by geodesic search.
  \Comment{\textcolor{blue}{\eqref{eq:shooting}}}

  \Statex \hspace{\algorithmicindent}\textcolor{blue}{\textit{3. Candidate validation}}
  \State Evaluate each $\boldsymbol{\zeta}\in\eta$ using one unperturbed and
  $Q$ action-error-perturbed complete rollouts with common random numbers.
  \State $\eta\gets
  \{\boldsymbol{\zeta}\in\eta:
  \mathbb I_{\mathrm{succ}}
  (\tau^{(0)}(\omega,\boldsymbol{\zeta}))=1\}$.
  \State Compute $J(\boldsymbol{\zeta})$ and
  $\mathrm{SR}(\boldsymbol{\zeta}\mid\omega)$ for all
  $\boldsymbol{\zeta}\in\eta$.
  \Comment{\textcolor{blue}{\eqref{eq:unperturbedquality}}}

  \If{$\mathbb I_{\mathrm{succ}}(\tau_{\mathrm{org}})=1$}
    \State Replay $\boldsymbol{\zeta}_{\mathrm{org}}$ and obtain
    $J_{\mathrm{org}},\mathrm{SR}_{\mathrm{org}}$.
    \State $\eta_{\mathrm{elig}}\gets
    \{\boldsymbol{\zeta}\in\eta:
    J(\boldsymbol{\zeta})<J_{\mathrm{org}},\
    \mathrm{SR}(\boldsymbol{\zeta}\mid\omega)
    \geq\mathrm{SR}_{\mathrm{org}}\}$.
  \Else
    \State $\eta_{\mathrm{elig}}\gets
    \{\boldsymbol{\zeta}\in\eta:
    \mathrm{SR}(\boldsymbol{\zeta}\mid\omega)
    \geq\delta_{\mathrm{SR}}\}$.
  \EndIf

  \Statex \hspace{\algorithmicindent}\textcolor{blue}{\textit{4. Demonstration selection}}
  \If{$\eta_{\mathrm{elig}}\neq\varnothing$}
    \State $\boldsymbol{\zeta}^{*}\in
    \arg\min_{\boldsymbol{\zeta}\in\eta_{\mathrm{elig}}}
    J(\boldsymbol{\zeta})$.
    \Comment{ties: \S\ref{sec:selection}}
    \State Retain every successful
    $\tau^{(q)}(\omega,\boldsymbol{\zeta}^{*})$,
    $q=0,\ldots,Q$.
  \ElsIf{$\mathbb I_{\mathrm{succ}}(\tau_{\mathrm{org}})=1$}
    \State Retain $\tau_{\mathrm{org}}$.
  \EndIf

\EndFor

\State Build $\mathcal D_{*}$ from all retained successful rollouts.
\State \Return $\mathcal D_{*}$
\end{algorithmic}
\end{algorithm}

\begin{figure*}[htbp]
  \centering
  \IfFileExists{figures/objects_catching.pdf}{    \includegraphics[width=0.96\textwidth]{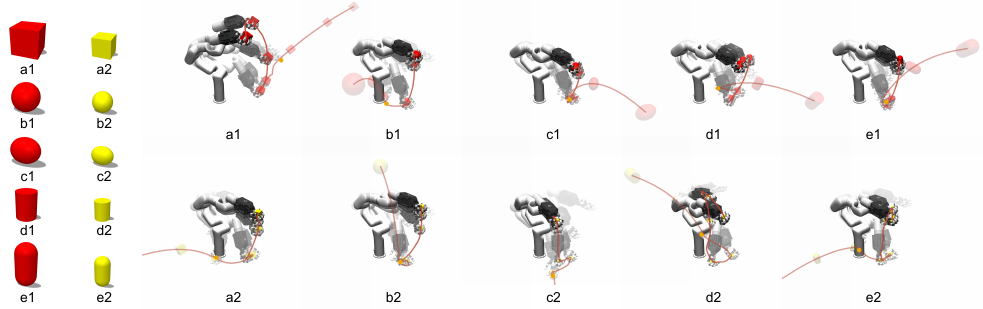}  }{    \fbox{\parbox[c][2.20in][c]{0.94\textwidth}{\centering\footnotesize
    Upload \texttt{objects\_catching.pdf} to the
    \texttt{figures/} directory.}}  }
  \vspace{-2.5mm} 
  \caption{Ten object variants and their corresponding impact-mitigating catches. The five families are box, sphere, ellipsoid, cylinder, and capsule, each shown in large red (1) and small yellow (2) variants.}
  \label{fig:objectcatches}
\vspace{-1mm} 
\end{figure*}

\begin{figure*}[htbp]
  \centering
  \IfFileExists{figures/catch_process.pdf}{    \includegraphics[width=0.98\textwidth]{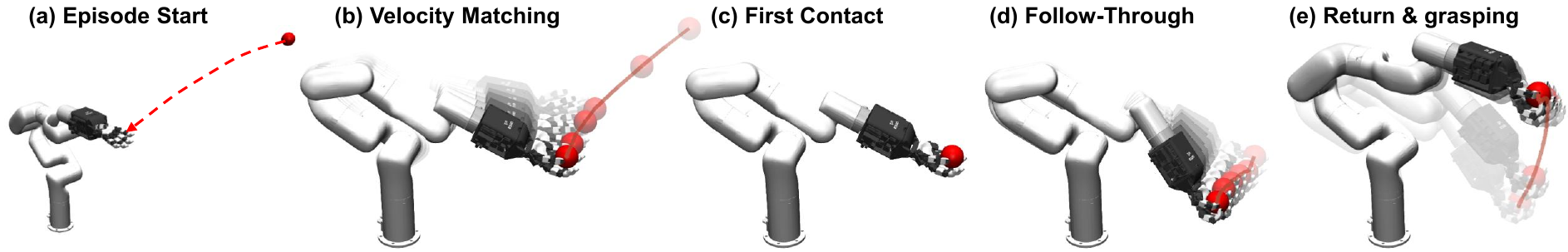}  }{    \fbox{\parbox[c][1.65in][c]{0.95\textwidth}{\centering\footnotesize
    Upload \texttt{catch\_process.pdf} to the \texttt{figures/} directory.}}  }
  \vspace{-2mm}   \caption{Representative impact-mitigating catch by the RL teacher. Faded poses show arm--hand and
  object motion; the red curve traces the object center of mass.}
  \label{fig:catchprocess}
\vspace{-1mm} \end{figure*}

\section{Experiments}
\label{sec:experiments}

We evaluate impact mitigation, the effect of demonstration construction on
closed-loop performance, performance across task conditions, and the
contributions of individual components.

\subsection{Experimental Setup}
\label{sec:systemsetup}

\textbf{Simulation.}
We use a 7-DoF xArm7 and a 17-DoF ORCA Hand in MuJoCo, with all policies
running at 50~Hz. We randomize object family and size, mass (uniform on
$[0.04,0.06]$~kg), launch position and speed (approximately 2.8--5.0~m/s),
and initial arm--hand posture. These launch randomizations induce the absolute lateral target offset, defined as the object's lateral distance from the initial palm center when crossing its horizontal plane. Domain randomization during RL-teacher training and
evaluation of $\bar{\pi}$ and $\pi_{\theta}$ includes object-position
perturbations, observation and actuation delays, actuation noise and gain
variation, control-timing jitter, and contact-dynamics variation.
Figure~\ref{fig:objectcatches} illustrates interception and follow-through
across the ten shape--size variants.

\textbf{Policies and search.}
The privileged-state RL teacher is trained with PPO using catching,
grasp-retention, velocity-matching, peak-impact-force mitigation, and
follow-through objectives. IL students
$\bar{\pi}$ and final students $\pi_{\theta}$ use conditional flow
matching~\cite{lipman2023flow,chisari2025pointflow} with histories of
deployable observations and previous actions; object position is their only
object-state input. Each predicts an eight-step action chunk, executes its
first action, and replans at every control step.
The datasets $\mathcal D_{\mathcal M}$ and $\mathcal D_{\mathcal E}$ each
contain 20,000 successful RL rollouts; $\mathcal D_{\mathcal R}$ contains
20,000 mixed rollouts. Search uses $\mathcal Z=[-3,3]^8$, $K=4$ shooting
directions, and radius $r=0.75$, producing nine candidates including the
decoded search center. We set $H_{\mathrm{OS}}=8$, $Q=16$, repair threshold
$\delta_{\mathrm{SR}}=0.75$, and error amplification $\kappa_\epsilon=1.5$.

\textbf{Baselines.}
Naive-IL, Normal-IL, and Ours-IL are trained on demonstrations from
success-only filtering, unperturbed-only validation, and Outcome-Sensitive
Motion Search, respectively. Unperturbed-only uses the same candidate sets
as our method but selects the lowest-cost candidate that succeeds without
modeled student action errors. All final students share the same interface,
architecture, action-chunk length, and optimization protocol. Ours-IL yields 99,328 successful demonstrations after perturbation. To control for dataset size, Naive-IL and Normal-IL are extended to the same count using additional successful RL rollouts and mixed rollouts, respectively.

\textbf{Evaluation.}
Task success follows Section~\ref{sec:problem}, with
$\delta_v=5.0$~m/s, $\delta_f=80$~N, $\delta_{\ell}=0.2$~m,
$\Delta t_{\mathrm{eval}}=0.20$~s, and $\Delta t_{\mathrm{grasp}}=1.0$~s.
To decouple catching performance from impact mitigation, we additionally report catch completion, which retains all success requirements except the three impact-mitigation conditions,
$v_{\mathrm{rel}}(t_c) \leq \delta_v$,
$f_{\mathrm{max}} \leq \delta_f$, and
$\ell_{\mathrm{max}} \geq \delta_\ell$. Impact metrics are reported only for catch completions. Each final-student
variant is trained with three seeds and evaluated on the same 3,000
held-out task conditions. Unless noted otherwise, the text reports seed
means; Tables~\ref{tab:constructionpolicy} and \ref{tab:errorablation}
report final-student means $\pm$ standard deviations across seeds.

\begin{figure*}[htbp]
  \centering
  \includegraphics[width=\textwidth]{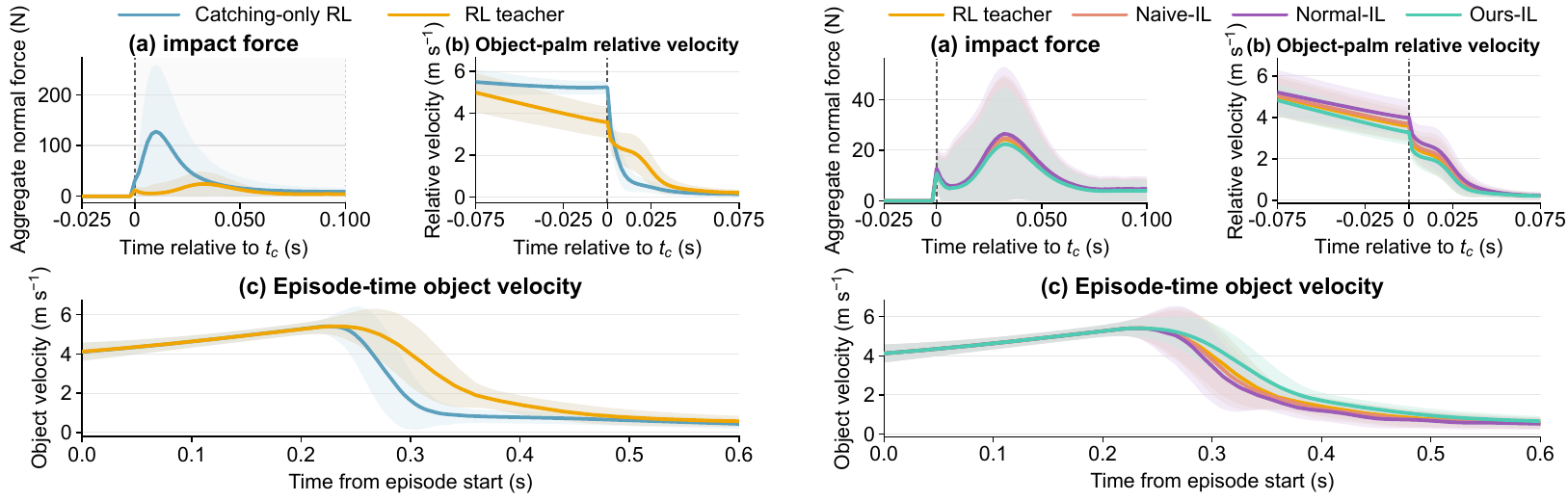}
  \vspace{-7mm}   \caption{Impact-mitigation profiles. Left: catching-only RL vs. the impact-mitigating RL teacher; right: the RL teacher vs. final students. (a) Aggregate normal force, (b) object–palm relative velocity, and (c) object velocity along the pre-contact incoming direction; (a,b) are aligned at $t_c$ (dashed line). Shaded bands denote the mean $\pm$ one sample standard deviation across rollouts at each time step.}
  \label{fig:ppopolicycomparison}
\vspace{0mm} \end{figure*}

\subsection{Catching with Impact Mitigation}
\label{sec:taskvalidation}

To distinguish catch completion from impact mitigation, we train a catching-only RL baseline with the teacher architecture and budget but without velocity-matching, impact-force, or follow-through objectives. Despite similar catch completion, the impact-mitigating teacher improves task success by 84.8 percentage points and reduces peak force by 74.5\% (Table~I), confirming that completion alone does not establish impact mitigation. Figure~\ref{fig:catchprocess} shows a
representative teacher catch; the profiles in
Fig.~\ref{fig:ppopolicycomparison} (left) illustrate the lower impact peak
and more gradual deceleration during follow-through.

\begin{table}[htbp]
\caption{Catching performance with and without impact mitigation.}
\vspace{-2mm}
\label{tab:impacttask}
\vspace{0mm} \centering
\scriptsize
\setlength{\tabcolsep}{3.0pt}
\begin{tabular}{@{}l c c c c c@{}}
\toprule
Controller & Catch & Task & $v_{\mathrm{rel}}$ & $f_{\mathrm{max}}$ &
$\ell_{\mathrm{max}}$\\
 & compl. & succ. & (m/s) & (N) & (m)\\
\midrule
Catching-only RL & 92.5\% & 0.4\% & 5.24 &
160.1 & 0.07\\
RL teacher & 90.7\% & 85.2\% & 3.60 &
40.8 & 0.42\\
\bottomrule
\end{tabular}
\vspace{-3mm} \end{table}

\begin{table*}[htbp]
\caption{Demonstration construction and policy performance.}
\label{tab:constructionpolicy}
\vspace{-2mm} \centering
\scriptsize
\setlength{\tabcolsep}{1.8pt}
\renewcommand{\arraystretch}{1.10}
\begin{tabular}{@{}>{\raggedright\arraybackslash}p{0.205\textwidth}
>{\centering\arraybackslash}p{0.083\textwidth}
>{\centering\arraybackslash}p{0.083\textwidth}
>{\centering\arraybackslash}p{0.088\textwidth}
>{\centering\arraybackslash}p{0.082\textwidth}
>{\centering\arraybackslash}p{0.120\textwidth}
>{\centering\arraybackslash}p{0.120\textwidth}
>{\centering\arraybackslash}p{0.105\textwidth}@{}}
\toprule
& \multicolumn{4}{c}{Demonstration construction}
& \multicolumn{3}{c}{Closed-loop policy performance}\\
\cmidrule(lr){2-5}\cmidrule(l){6-8}
Construction / resulting policy
& \shortstack{Demo.\\coverage}
& \shortstack{Repair\\rate}
& \shortstack{Refinement\\rate}
& \shortstack{Unperturbed\\$f_{\mathrm{max}}$ (N)}
& \shortstack{Task\\success}
& \shortstack{Catch\\compl.}
& \shortstack{Policy\\$f_{\mathrm{max}}$ (N)}\\
\midrule
RL teacher (original)
& -- & -- & -- & --
& 85.2\% & 90.7\% & 40.8\\
Success-only filtering / Naive-IL
& 85.2\% & 0\% & 0\% & 40.6
& 83.8\% $\pm$ 0.6\% & 89.9\% $\pm$ 0.5\% & 41.1 $\pm$ 1.0\\
Unperturbed-only / Normal-IL
& 94.9\% & 65.5\% & 84.2\% & 35.9
& 81.7\% $\pm$ 0.7\% & 88.0\% $\pm$ 0.9\% & 44.2 $\pm$ 1.8\\
Our method / Ours-IL
& 91.8\% & 44.3\% & 59.8\% & 36.5
& 90.3\% $\pm$ 0.8\% & 93.6\% $\pm$ 0.4\% & 37.3 $\pm$ 1.2\\
\bottomrule
\end{tabular}
\vspace{-1mm} \end{table*}

\subsection{Demonstration Construction and Policy Performance}
\label{sec:datacomparison}
\label{sec:policycomparison}

Table~\ref{tab:constructionpolicy} relates demonstration construction to
closed-loop policy performance. Coverage is the fraction of target task
conditions represented by successful unperturbed rollouts. Repair rate is
the fraction of failed RL originals converted to successful unperturbed
rollouts; refinement rate is the fraction of successful originals replaced
by eligible candidates.

Ours-IL achieves 90.3\% task success, exceeding the RL teacher by
5.1 percentage points and Naive-IL by 6.5 points, while reducing peak impact
force by 8.6\% relative to the teacher. In contrast, unperturbed-only
validation gives the highest coverage and lowest unperturbed
$f_{\mathrm{max}}$, yet Normal-IL trails Naive-IL by 2.1 points. Our method
has lower coverage than unperturbed-only validation but better closed-loop
performance, showing that
unperturbed demonstration quality alone does not determine imitation
performance. Figure~\ref{fig:ppopolicycomparison} (right) further shows
Ours-IL's lower pre-contact object--palm relative speed, lowest impact
peak, and more gradual deceleration during follow-through.

\Needspace{8\baselineskip}
\subsection{Performance Across Task Conditions}
\label{sec:difficultygeneralization}

\begin{figure}[htbp]
  \centering
  \IfFileExists{figures/representative_scenarios.pdf}{    \includegraphics[width=\columnwidth]{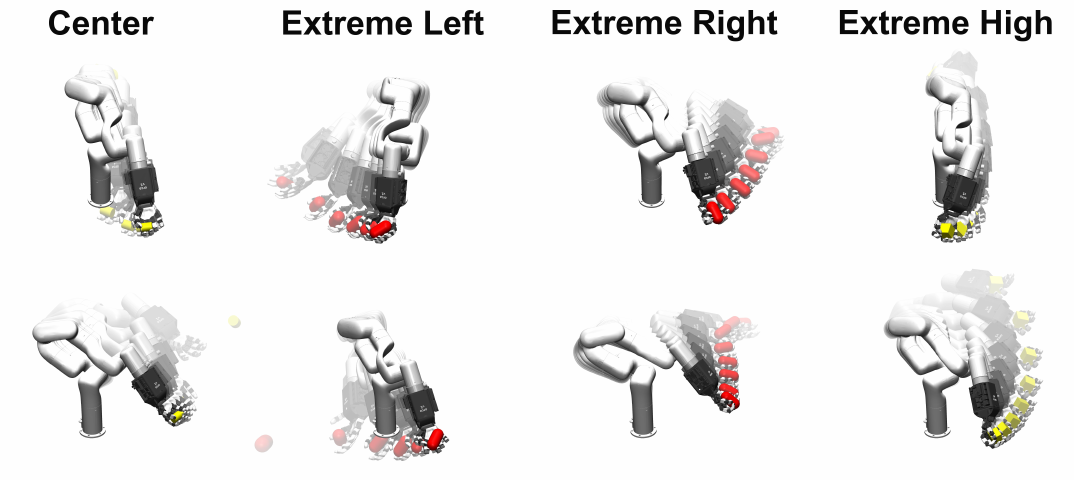}  }{    \fbox{\parbox[c][1.48in][c]{0.94\columnwidth}{\centering\footnotesize
    Upload \texttt{representative\_scenarios.pdf} to the
    \texttt{figures/} directory.}}  }
  \vspace{-6.5mm}   \caption{Impact-mitigating RL teacher catches for a central and three extreme
  launch directions. Each column shows two views of one rollout; faded
  arm--hand poses trace catching and follow-through.}
  \label{fig:representativescenarios}
\vspace{-3mm} \end{figure}

\begin{figure}[htbp]
  \centering
  \includegraphics[width=\columnwidth]{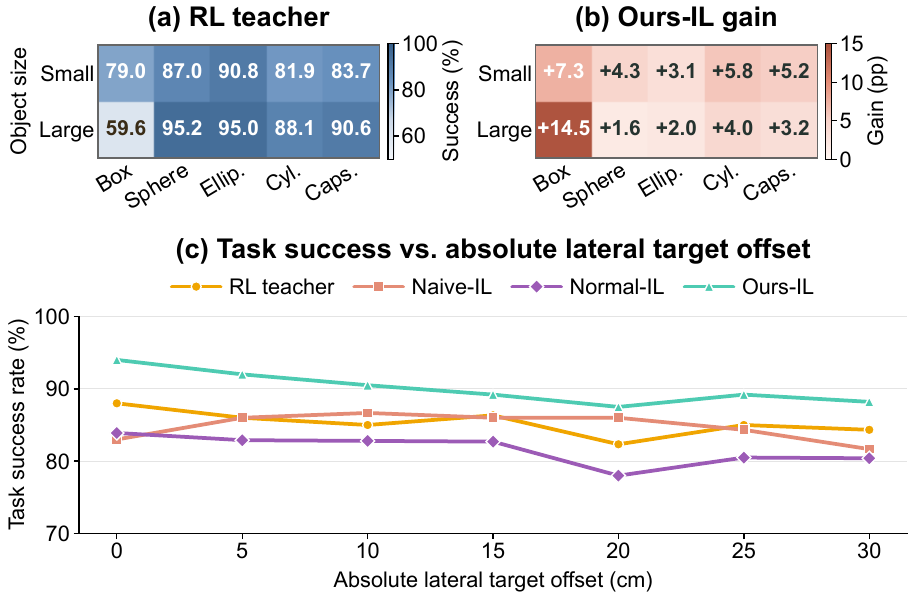}
  \vspace{-5.5mm}   \caption{Performance across task conditions. (a) RL teacher task success
  rate by object family and size; (b) Ours-IL gain over the RL teacher (percentage
  points); (c) policy task success rate versus absolute lateral target offset.}
  \label{fig:taskdifficulty}
\vspace{-2mm} \end{figure}

Figure~\ref{fig:taskdifficulty} compares performance across object shape,
size, and absolute lateral target offset. Ours-IL improves all ten shape--size combinations, with the largest gain (+14.5 percentage points) on the large box, and maintains the highest success across lateral target offsets
Figure~\ref{fig:representativescenarios} shows representative teacher
catches for a central and three extreme launch directions.

\subsection{Ablation Studies}
\label{sec:constructionanalysis}
\label{sec:ablations}

\textbf{Refinement and repair.}
At matched dataset size, success-only filtering yields 83.8\% task success
and 41.1~N peak impact force. Refinement-only modifies successful RL
originals without repairing failures, reducing force to 36.2~N with
83.2\% task success. Repair-only repairs failed conditions while retaining
successful originals unchanged, achieving 88.5\% success and 42.7~N.
Combining both yields the highest success rate (90.3\%) with 37.3~N.
Refinement primarily improves impact mitigation, whereas repair provides
most of the success gain.

\textbf{Candidate generation.}
We match candidate-set size, maximum action-space deviation from the
original arm-action snippet, and dataset size. Repair/refinement/student-success rates increase from 5.6/8.3/83.4\% with direct perturbations to 25.7/33.1/86.8\% with an unconditional manifold, 39.6/51.9/88.5\% with conditional Euclidean search, and 44.3/59.8/90.3\% with the full geodesic search. These comparisons support the
contributions of task conditioning and geodesic search.

\textbf{Action-error modeling.}
Table~\ref{tab:errorablation} compares unperturbed-only validation with
three models calibrated from $\bar{\pi}$, using identical candidate sets
and matched dataset size. Global i.i.d. Gaussian ignores phase and temporal
correlation; phase-wise independent retains phase conditioning but removes
temporal correlation; phase-wise correlated is the full model in
Eq.~\eqref{eq:errormodel}. 
Moment error measures mismatch in phase-wise action-error distribution
moments; lag-one error measures mismatch in lag-one temporal
autocorrelation.

\begin{table}[htbp]
\caption{Comparison of action-error models.}
\label{tab:errorablation}
\vspace{-2mm} \centering
\scriptsize
\setlength{\tabcolsep}{1.2pt}
\renewcommand{\arraystretch}{1.05}
\begin{tabular}{@{}>{\raggedright\arraybackslash}p{0.29\columnwidth}
>{\centering\arraybackslash}p{0.13\columnwidth}
>{\centering\arraybackslash}p{0.13\columnwidth}
>{\centering\arraybackslash}p{0.13\columnwidth}
>{\centering\arraybackslash}p{0.21\columnwidth}@{}}
\toprule
Validation model & \shortstack{Moment\\err. $\downarrow$} &
\shortstack{Lag-1\\err. $\downarrow$} & \shortstack{Repair\\rate} &
\shortstack{$\pi_{\theta}$ task\\succ.}\\
\midrule
Unperturbed-only & -- & -- & 65.5\% &
81.7\% $\pm$ 0.7\%\\
Global i.i.d. Gaussian & 0.289 & 0.142 & 61.6\% &
82.2\% $\pm$ 1.3\%\\
Phase-wise independent & 0.012 & 0.132 &
48.6\% & 86.8\% $\pm$ 0.7\%\\
Phase-wise correlated & 0.038 & 0.041 &
44.3\% & 90.3\% $\pm$ 0.8\%\\
\bottomrule
\end{tabular}
\vspace{-2mm} \end{table}

Phase conditioning improves moment matching, while temporal correlation
modeling gives the lowest lag-one error and highest student task success,
despite a lower repair rate.

\textbf{Limitations.}
Our search cannot repair failures outside the support of the learned OSW
motion manifold, including those requiring different hand control or
post-contact recovery. Candidate selection depends on simulated contact
dynamics and the action-error model calibrated from $\bar{\pi}$ for
$\pi_{\theta}$. Robustness to real perception and contact mismatch remains
unverified without physical arm--hand experiments.\par

\section{Conclusion}
Outcome-Sensitive Motion Search constructs demonstrations robust to modeled execution errors of $\pi_{\theta}$
from mixed RL experience through OSW motion manifold search and complete-rollout
validation under the action-error model calibrated from $\bar{\pi}$. In simulation, the resulting
observation-restricted final student $\pi_{\theta}$ surpasses the fixed privileged teacher while
reducing impact force, demonstrating the importance of both demonstration quality
and robustness to modeled execution errors of $\pi_{\theta}$ for impact-aware catching. Future work will evaluate
physical arm--hand deployment and broader throwing conditions.

\bibliographystyle{IEEEtran}
\bibliography{references}

@INPROCEEDINGS{gold2022mpc,
  author={Gold, Tobias and Römer, Ralf and Völz, Andreas and Graichen, Knut},
  booktitle={2022 American Control Conference (ACC)}, 
  title={Catching Objects with a Robot Arm using Model Predictive Control}, 
  year={2022},
  volume={},
  number={},
  pages={1915-1920},
  doi={10.23919/ACC53348.2022.9867380}}

@INPROCEEDINGS{bauml2010catching,
  author={Bäuml, Berthold and Wimböck, Thomas and Hirzinger, Gerd},
  booktitle={2010 IEEE/RSJ International Conference on Intelligent Robots and Systems}, 
  title={Kinematically optimal catching a flying ball with a hand-arm-system}, 
  year={2010},
  volume={},
  number={},
  pages={2592-2599},
  doi={10.1109/IROS.2010.5651175}}

@ARTICLE{salehian2016softly,
  author={Salehian, Seyed Sina Mirrazavi and Khoramshahi, Mahdi and Billard, Aude},
  journal={IEEE Transactions on Robotics}, 
  title={A Dynamical System Approach for Softly Catching a Flying Object: Theory and Experiment}, 
  year={2016},
  volume={32},
  number={2},
  pages={462-471},
  doi={10.1109/TRO.2016.2536749}}

@INPROCEEDINGS{zhao2023impactfriendly,
  author={Zhao, Jianzhuang and Lahr, Gustavo J. G. and Tassi, Francesco and Santopaolo, Alessandro and De Momi, Elena and Ajoudani, Arash},
  booktitle={2023 IEEE/RSJ International Conference on Intelligent Robots and Systems (IROS)}, 
  title={Impact-Friendly Object Catching at Non-Zero Velocity Based on Combined Optimization and Learning}, 
  year={2023},
  volume={},
  number={},
  pages={4428-4435},
  doi={10.1109/IROS55552.2023.10341600}}

@ARTICLE{yan2024impact,
  author={Yan, Lei and Stouraitis, Theodoros and Moura, João and Xu, Wenfu and Gienger, Michael and Vijayakumar, Sethu},
  journal={IEEE Transactions on Robotics}, 
  title={Impact-Aware Bimanual Catching of Large-Momentum Objects}, 
  year={2024},
  volume={40},
  number={},
  pages={2543-2563},
  doi={10.1109/TRO.2024.3381551}}

@misc{lan2024dexcatch,
      title={DexCatch: Learning to Catch Arbitrary Objects with Dexterous Hands}, 
      author={Fengbo Lan and Shengjie Wang and Yunzhe Zhang and Haotian Xu and Oluwatosin Oseni and Ziye Zhang and Yang Gao and Tao Zhang},
      year={2024},
      eprint={2310.08809},
      archivePrefix={arXiv},
      primaryClass={cs.RO},
      url={https://arxiv.org/abs/2310.08809}, 
}

@misc{zhang2025catchit,
      title={Catch It! Learning to Catch in Flight with Mobile Dexterous Hands}, 
      author={Yuanhang Zhang and Tianhai Liang and Zhenyang Chen and Yanjie Ze and Huazhe Xu},
      year={2024},
      eprint={2409.10319},
      archivePrefix={arXiv},
      primaryClass={cs.RO},
      url={https://arxiv.org/abs/2409.10319}, 
}

@INPROCEEDINGS{hu2023modular,
  author={Hu, Wenbin and Acero, Fernando and Triantafyllidis, Eleftherios and Liu, Zhaocheng and Li, Zhibin},
  booktitle={2023 IEEE/RSJ International Conference on Intelligent Robots and Systems (IROS)}, 
  title={Modular Neural Network Policies for Learning In-Flight Object Catching with a Robot Hand-Arm System}, 
  year={2023},
  volume={},
  number={},
  pages={944-951},
  doi={10.1109/IROS55552.2023.10341463}}

@INPROCEEDINGS{lampariello2011trajectory,
  author={Lampariello, Roberto and Nguyen-Tuong, Duy and Castellini, Claudio and Hirzinger, Gerd and Peters, Jan},
  booktitle={2011 IEEE International Conference on Robotics and Automation}, 
  title={Trajectory planning for optimal robot catching in real-time}, 
  year={2011},
  volume={},
  number={},
  pages={3719-3726},
  doi={10.1109/ICRA.2011.5980114}}

@misc{ross2011dagger,
      title={A Reduction of Imitation Learning and Structured Prediction to No-Regret Online Learning}, 
      author={Stephane Ross and Geoffrey J. Gordon and J. Andrew Bagnell},
      year={2011},
      eprint={1011.0686},
      archivePrefix={arXiv},
      primaryClass={cs.LG},
      url={https://arxiv.org/abs/1011.0686}, 
}

@misc{laskey2017dart,
      title={DART: Noise Injection for Robust Imitation Learning}, 
      author={Michael Laskey and Jonathan Lee and Roy Fox and Anca Dragan and Ken Goldberg},
      year={2017},
      eprint={1703.09327},
      archivePrefix={arXiv},
      primaryClass={cs.LG},
      url={https://arxiv.org/abs/1703.09327}, 
}

@misc{ke2024ccil,
      title={CCIL: Continuity-based Data Augmentation for Corrective Imitation Learning}, 
      author={Liyiming Ke and Yunchu Zhang and Abhay Deshpande and Siddhartha Srinivasa and Abhishek Gupta},
      year={2024},
      eprint={2310.12972},
      archivePrefix={arXiv},
      primaryClass={cs.RO},
      url={https://arxiv.org/abs/2310.12972}, 
}

@INPROCEEDINGS{hoque2024intervengen,
  author={Hoque, Ryan and Mandlekar, Ajay and Garrett, Caelan and Goldberg, Ken and Fox, Dieter},
  booktitle={2024 IEEE/RSJ International Conference on Intelligent Robots and Systems (IROS)}, 
  title={IntervenGen: Interventional Data Generation for Robust and Data-Efficient Robot Imitation Learning}, 
  year={2024},
  volume={},
  number={},
  pages={2840-2846},
  doi={10.1109/IROS58592.2024.10801523}}

@misc{mandlekar2022offline,
      title={What Matters in Learning from Offline Human Demonstrations for Robot Manipulation}, 
      author={Ajay Mandlekar and Danfei Xu and Josiah Wong and Soroush Nasiriany and Chen Wang and Rohun Kulkarni and Li Fei-Fei and Silvio Savarese and Yuke Zhu and Roberto Martín-Martín},
      year={2021},
      eprint={2108.03298},
      archivePrefix={arXiv},
      primaryClass={cs.RO},
      url={https://arxiv.org/abs/2108.03298}, 
}

@misc{xu2022dwbc,
      title={Discriminator-Weighted Offline Imitation Learning from Suboptimal Demonstrations}, 
      author={Haoran Xu and Xianyuan Zhan and Honglei Yin and Huiling Qin},
      year={2022},
      eprint={2207.10050},
      archivePrefix={arXiv},
      primaryClass={cs.LG},
      url={https://arxiv.org/abs/2207.10050}, 
}

@inproceedings{mandlekar2023mimicgen,
    title={MimicGen: A Data Generation System for Scalable Robot Learning using Human Demonstrations},
    author={Mandlekar, Ajay and Nasiriany, Soroush and Wen, Bowen and Akinola, Iretiayo and Narang, Yashraj and Fan, Linxi and Zhu, Yuke and Fox, Dieter},
    booktitle={7th Annual Conference on Robot Learning},
    year={2023}
}

@INPROCEEDINGS{hertel2021learning,
  author={Hertel, Brendan and Ahmadzadeh, S. Reza},
  booktitle={2021 IEEE/RSJ International Conference on Intelligent Robots and Systems (IROS)}, 
  title={Learning from Successful and Failed Demonstrations via Optimization}, 
  year={2021},
  volume={},
  number={},
  pages={7807-7812},
  doi={10.1109/IROS51168.2021.9636679}}

@inproceedings{kim2022demodice,
  title     = {DemoDICE: Offline Imitation Learning with Supplementary Imperfect Demonstrations},
  author    = {Geon-Hyeong Kim and Seokin Seo and Jongmin Lee and Wonseok Jeon and HyeongJoo Hwang and Hongseok Yang and Kee-Eung Kim},
  booktitle = {International Conference on Learning Representations},
  year      = {2022}
}

@INPROCEEDINGS{chen2025s2i,
  author={Chen, Jingjing and Fang, Hongjie and Fang, Hao-Shu and Lu, Cewu},
  booktitle={2025 IEEE International Conference on Robotics and Automation (ICRA)}, 
  title={Towards Effective Utilization of Mixed-Quality Demonstrations in Robotic Manipulation via Segment-Level Selection and Optimization}, 
  year={2025},
  volume={},
  number={},
  pages={16884-16891},
  doi={10.1109/ICRA55743.2025.11128787}}

@misc{lipman2023flow,
      title={Flow Matching for Generative Modeling}, 
      author={Yaron Lipman and Ricky T. Q. Chen and Heli Ben-Hamu and Maximilian Nickel and Matt Le},
      year={2023},
      eprint={2210.02747},
      archivePrefix={arXiv},
      primaryClass={cs.LG},
      url={https://arxiv.org/abs/2210.02747}, 
}

@misc{chisari2025pointflow,
      title={Learning Robotic Manipulation Policies from Point Clouds with Conditional Flow Matching}, 
      author={Eugenio Chisari and Nick Heppert and Max Argus and Tim Welschehold and Thomas Brox and Abhinav Valada},
      year={2024},
      eprint={2409.07343},
      archivePrefix={arXiv},
      primaryClass={cs.RO},
      url={https://arxiv.org/abs/2409.07343}, 
}

@article{tassi2026imacatcher,
   title={IMA-catcher: An IMpact-aware nonprehensile catching framework based on combined optimization and learning},
   volume={45},
   ISSN={1741-3176},
   url={http://dx.doi.org/10.1177/02783649251345851},
   DOI={10.1177/02783649251345851},
   number={1},
   journal={The International Journal of Robotics Research},
   publisher={SAGE Publications},
   author={Tassi, Francesco and Zhao, Jianzhuang and Lahr, Gustavo JG and Gava, Luna and Monforte, Marco and Glover, Arren and Bartolozzi, Chiara and Ajoudani, Arash},
   year={2025},
   month={June}, pages={100--127} }

@INPROCEEDINGS{pei2026kinodynamic,
  author={Pei, Guorui and Zhang, Mengshi and Chen, Xi and Wu, Jinsong and Qi, Jiaming and Zhou, Peng},
  booktitle={2026 7th International Conference on Mechatronics Technology and Intelligent Manufacturing (ICMTIM)}, 
  title={Learning a Kinodynamic Trajectory Manifold for Impact-Aware Compliant Catching of Fast-Moving Objects}, 
  year={2026},
  volume={},
  number={},
  pages={156-160},
  doi={10.1109/ICMTIM69588.2026.11525755}}

@article{wang2026world,
  title={World models for robotic manipulation: A survey},
  author={Wang, Fangyuan and Wang, Ziyuan and Pei, Guorui and Zhang, Mengshi and Liang, Canxi and Hu, Jun and Li, Zhongxuan and Wu, Jinsong and Han, Ning and Zhang, Zeqing and others},
  journal={SmartBot},
  pages={e70053},
  year={2026},
  publisher={Wiley Online Library}
}

\end{document}